\documentclass[10pt, a4paper]{article}
\usepackage{lrec}
\usepackage{multibib}
\newcites{languageresource}{Language Resources}
\usepackage{graphicx}
\usepackage{tabularx}
\usepackage{booktabs}
\usepackage{soul}
\usepackage[utf8]{inputenc}

\usepackage{hyperref}
\usepackage{cleveref}
\usepackage{xstring}
\usepackage{xspace}
\usepackage[autostyle]{csquotes}
\usepackage[autolanguage]{numprint}
\usepackage[english]{isodate}

\usepackage{xcolor}

\newcommand{\timex}{\textsc{timex3}\xspace}
\newcommand{\Name}{HeidelTime\textsubscript{ext}\xspace}

\title{I still have Time(s): Extending HeidelTime for German Texts}

\name{Andy Lücking, Manuel Stoeckel, Giuseppe Abrami, Alexander Mehler} 

\address{Goethe University Frankfurt, Text Technology Lab \\
         Robert-Mayer-Str. 10, D-60325 Frankfurt am Main, Germany \\
         \{luecking, m.stoeckel, abrami, mehler\}@em.uni-frankfurt.de\\}

\abstract{
HeidelTime is one of the most widespread and successful 
tools for detecting temporal expressions in texts. 
Since HeidelTime's pattern matching system is based on regular expression, it can be extended in a convenient way.
We present such an extension for the German resources of HeidelTime: \Name.
The extension has been brought about by means of observing false negatives within real world texts and various time banks. 
The gain in coverage is \numprint[\%]{2.7} or \numprint[\%]{8.5}, depending on the admitted degree of potential overgeneralization.
We describe the development of \Name, its evaluation on text samples from various genres, and share some linguistic observations. 
\Name can be obtained from \url{https://github.com/texttechnologylab/heideltime}.
\\ \newline \Keywords{HeidelTime, German, TimeBanks, BIOfid} 
}

\begin{document}

\maketitleabstract

\section{Motivation}
\label{sec:motivation}

Biodiversity literature exhibits frequent references to times. 
Hence, processing biodiversity texts, as is done in the context of specialized information services \cite{Driller2020c}, involves recognizing temponyms, that is, temporal expressions which have -- possibly aided by context information -- a unique interpretation on a time line, like toponyms do in terms of geo-spatial entities \cite{Kuzey:et:al:2016}.
To this end, HeidelTime \cite{HeidelTime:2016,Stroetgen:Gertz:2013} is used within the BIOfid project (\url{www.biofid.de}) to detect mentions of time-denoting expressions in mainly German biological texts from the 19th century until today -- see \cite{BIOfid-LRE:2021} for an overview.
These texts, however, contain temponyms that are outside of the extension of HeidelTime's rule system and include spelling variants of mundane temponyms like \textit{Herbste} \enquote*{fall}\footnote{E.g., \url{https://sammlungen.ub.uni-frankfurt.de/3728500} and \url{https://sammlungen.ub.uni-frankfurt.de/4497114}}, duration-forming constructions such as \textit{in den letzten beiden Jahren} \enquote*{in the past two years}\footnote{E.g., \url{https://sammlungen.ub.uni-frankfurt.de/3721623}}, and set-forming constructions such as \textit{in einem zweijährigen Turnus} \enquote*{on a biennial basis}\footnote{E.g., \url{https://sammlungen.ub.uni-frankfurt.de/3748527}}.
Based on such missing patterns of temporal expressions, among others, we extend the German component of HeidelTime (\cref{sec:procedure}).
We want to emphasize that HeidelTime is designed in such a way that such extensions can be implemented in a comparatively easy manner. 
The gain in coverage in assessed on various kinds of corpora (\cref{sec:evaluation}). 
A manual inspection of the novel instances found by \Name brought two overgeneralizing rules to the light, however.
Depending on how much potential overgeneralization is admitted, the relative coverage improvement of \Name is \numprint[\%]{2.7} (excluding one source of overgeneralization) or \numprint[\%]{8.5} (including the source of overgeneralization).
The HeidelTime extension -- called \Name\ -- is available at \url{https://github.com/texttechnologylab/heideltime}.

\section{Procedure}
\label{sec:procedure}

HeidelTime is developed to detect a certain kind of temponyms, namely temponyms that are categorized as \timex expressions \cite{ISO24617-1}.
\timex expressions comprise dates (e.g., \textit{10 January, 1999}), times (e.g., \textit{12 o'clock}), durations (e.g., \textit{two weeks}), sets (e.g., \textit{every year}).
%
To extend HeidelTime, we first looked for \timex patterns which are not yet covered.

\subsection{Manual Exploration}
\label{subsec:manual-exploration}

In the context of BIOfid, we manually sampled sentences from the current BIOfid corpus and processed the sample with HeidelTime.
We then inspected the outcome, focusing on false negatives, that is, \timex expressions that have not been detected.
Generalizing over these false negatives, missing patterns or expressions have been identified.
In contexts like natural language processing in biodiversity or the humanities it is important to cover all these instances irrespective of their frequency (we come back to matters of frequency as part of the evaluation in Sec.~\ref{sec:evaluation}).
In general, false negatives fall in one of four classes:
\begin{itemize}
    \item Spelling variants. This rather trivial class pertains to typographic variation. In German, the dative case of masculine singular nouns can be marked by the suffix \textit{-e}, as in \textit{dem Herbste} \enquote*{the-DAT fall}.
    This spelling variation, however, is a bit old-fashioned, so that it frequently occurs in the older texts from the BIOfid corpus, but less so in contemporary writings.
    Another example is punctuation in time expressions.
    Although using a dot instead of a colon to separate hour and minute -- as in \textit{21.30 Uhr} \enquote*{9.30 pm} -- does not comply with international standards \cite{ISO:8601}, it is nonetheless used in texts and for that reason should be detected. 
    
    \item Lexical extensions. Time spans are sometimes partitioned according to business or financial concerns, such as fiscal years.
    Corresponding nouns (e.g., \textit{Geschäftsjahr}) are a straightforward extension to HeidelTime's time units. 
    Further lexical extensions are due to temporal adjectives or adverbs.
    For instance, the modifier \textit{täglich} \enquote*{everyday} has the same meaning as the quantified noun phrase \textit{jeden Tag} \enquote*{every day}, but, in contrast to the noun phrase is not yet covered by HeidelTime's lexicon.
    %
    
    \item Compounds. German is well-known for its \enquote{tapeworm words} \cite{Twain:2016:awful}. 
    However, HeidelTime is not concerned with compounds, and presumably for a good reason: the risk of overgeneration is large.
    For instance, \textit{Jugendzeit} \enquote*{young days}/\enquote*{adolescence}/\enquote*{youth}, although ending on \textit{-zeit} \enquote*{-time}, is not a \timex expression but denotes a developmental stage.
    However, there are a couple of \enquote{well behaved} compounds. 
    We count compounds where the modifying noun is a known temponym among them, such as \textit{Winterzeit} \enquote*{wintertime} or  \textit{Sommermonate} \enquote*{months of summer}.
    Arguably, the head nouns do not contribute much in these cases so that, for instance, a combination of a season term and \textit{-zeit} \enquote*{-time} can be normalized to the value of the season term in a straightforward way. 
    Compounding also underlies the formation of temporal expressions of a set type. 
    An example is \textit{-basis}: \textit{Wochenbasis} \enquote*{on a weekly basis} means every week. 
    
    \item Rule extensions. Some of HeidelTime's rules are restricted to a certain class of expression.
    For instance, while quantifying over years is covered, quantifying over seasons is not. 
    For that reason, the duration denoting expression \textit{viele Winter} \enquote*{for many winters} is not recognized as such, but has become recognizable by adding corresponding duration rules. 
    A related observation can be made with regard to relative times. 
    There are rules that capture times such as \textit{letzten Freitag} \enquote*{last Friday}, but the synonymous expression \textit{vorheriger Freitag} \enquote*{previous Friday} had to be licensed by an additional rule.
\end{itemize}
All extensions are marked as such within the source files making up \Name.
It should be mentioned that not much emphasis is put on grammatical well-formedness or common usage: seldom or questionable compounds are recognized as well as phrases that lack morpho-syntactic agreement.
After all, HeidelTime, as \Name, is an annotator, not a grammar.

\subsection{Populating Negative Rules}
\label{subsec:negative-rules}

In writing regular expressions, care has to be taken to not to overgenerate.
To this end, HeidelTime employs so-called \emph{negative rules}, that is, rules which, when apply, remove their matched expressions from the output. 
To give an example: while season names are welcome temponyms, they can also figure as family names such as in \textit{Herr Sommer} \enquote*{Mister Summer}.
Such instances can be excluded by a negative rule that says that if a season term follows \textit{Herr} \enquote*{Mister} or \textit{Frau} \enquote*{Miss}, then remove it. 
We added such a negative rule.
However, one cannot stop here: 
the BIOfid example \textit{Assistent Sommer} \enquote*{assistant Summer} (file \url{https://sammlungen.ub.uni-frankfurt.de/3673151}) still circumvents our new negative rule.
Obviously, in addition to addressing particles, also profession terms can mark a season term as a proper name.
Therefore, we collected a list of profession terms from the German agency for employment and added them to \Name's pattern files.\footnote{To be more precise, we applied a string manipulation on the profession terms first: we removed subdomain classifications (for instance, distinguishing farmers for different agricultural sectors), and extended each entry into a separate masculine and feminine form. To avoid running into errors due to too long regular expressions, we restricted the list to single-word profession terms.}
Addressing particles and profession terms still fall short of capturing \textit{Ehepaar Sommer} \enquote*{the married couple Summer}, however.
This example is, of course, covered by \Name, but it illustrates that time recognition seems to be an open-ended task. 
For this reason, we followed a more dynamic approach and used BERT, a transformer-based language model trained for contextual embeddings of words \cite{devlin2019bert}.
BERT can be used like a cloze test: suggestions for a masked item can be obtained from left and right context information.
In this manner, we used the sentence containing \textit{Assistent Sommer} as input, masked the noun \textit{Assistent}, and collected the \numprint{30000} words (in fact, BERT also suggests non-word character sequences) which according to BERT fit best into the placeholder position.
We then removed all suggestions that are shorter than 4 characters and excluded fragmentary items (starting with \enquote{\#\#}).
Finally, we selected the first \numprint{5000} suggestions (using [much] more leads to regular expressions which are too long to handle for the system).
The BERT list provides an immediate benefit for the negative rule:  since it includes typical given names, full names ending on a season or weekday term are rightly excluded from \Name's temponym recognition.

\subsection{Harvesting Time Banks}
\label{subsec:harvest-timebanks}

To obtain indications of further extensions of HeidelTime, we looked at instances of expressions that have been marked as \timex expressions in several time banks. 
We extracted the content of \timex tags from the French TimeBank \cite{Bittar:Amsili:Denis:Danlos:2011} \citelanguageresource{FrenchTimeBank}, the Basque TimeBank \cite{EusTimeML:2020} \citelanguageresource{EusTimeML:resource}, and the MEANTIME newsreader corpus \cite{minard-etal-2016-meantime} \citelanguageresource{Newsreader:resource} (Dutch, English, Italian, Spanish). 
We then used \url{www.deepl.com} to translate the time expressions from different languages into German ones.
We fed the list into HeidelTime and inspected the outcome, most notably lines that lacked a HeidelTime tag.\footnote{Since the temporal expression-based input lacks sentential context, ambiguity due to conflicting pattern scopes arises. For instance, if a day term precedes a date, then both are taken to be a single time expression. If, however, a second day term follows, a conflict arises whether the first or the second day mention constitutes a time expression together with the date. Such ambiguities do not do harm to our approach, since the output is manually inspected anyway.}
This procedure resulted in 83 sample pattern which underlie the coverage extension of \Name. 
The examples have been chosen manually. 
False negatives which do not make up clear temponyms have been ignored.
This includes event-denotating expressions (e.g., \textit{l'heure de l'Europe et de la mondialisation} \enquote*{in the age of Europe and globalization} or \textit{temps de guerre contre le terrorisme} \enquote*{in times of the war on terrorism}), \enquote{vacuous} temporal expressions (such as \textit{any time}), and temporal modifiers which typically are used to modify not-temporal nouns (e.g., \textit{recent}).


\section{Evaluation}
\label{sec:evaluation}

The gain in coverage of HeidelTime and \Name (Sec.~\ref{sec:results}) is assessed and compared on various resources (Sec.~\ref{sec:evaluation-corpus}), which have been pre-processed as described in Sec.~\ref{sec:pipeline}.

\subsection{Evaluation Corpus}
\label{sec:evaluation-corpus}

Texts have been sampled from five different sources to balance potential effects of text type to the frequency of temporal expression use:
\begin{itemize}
\item 10 randomly collected protocols of the German Bundestag (\url{https://www.bundestag.de/services/opendata}).

\item 10 books from the \emph{German Text Archive} (DTA, \url{https://www.deutschestextarchiv.de}, namely Dickens, \textit{Weihnachtsabend} (1844), Fontane, \textit{Effi Briest} (1896), Goethe, \textit{Faust 1} (1808), von Humboldt, \textit{Kosmos}, vol. 1 (1845), Kafka, \textit{Die Verwandlung} (1915), Lessing, \textit{Nathan der Weise} (1779), Marx, \textit{Das Kapital}, vol. 1 (1867), Nietzsche, \textit{Homer und die klassische Philologie} (1869).

\item 10 randomly selected tests from the \emph{Zoologisch-Botanische Datenbank} (Zobobat, \url{https://www.zobodat.at}). The texts have been converted to plain text files from OCR PDFs and hence contain some token errors.

\item 766 articles from the \emph{Süddeutsche Zeitung} (SZ), collected within three TEI files with the following time stamps:  15.\,06.~1996, 04.\,09.~2002 and 07.\,12.~2013. 

\item \numprint{10000} randomly selected sentences from Wikipedia (WP) from the \emph{Leipzig Wortschatz}\footnote{\url{https://wortschatz.uni-leipzig.de/en/download}}, dump from the year 2012 \cite{Goldhahn:2012}.
\end{itemize}

The size of the samples is summarized in Tab.~\ref{tab:number-sentences}.

\begin{table}[htb]
    \centering
    \begin{tabular}{lll}
    \toprule
    Sample & \# sentences & \# tokens \\
    \midrule
        Bundestag &  \numprint{188768} & \numprint{3682370} \\
        DTA & \numprint{27687} & \numprint{810582} \\
        SZ & \numprint{18938} & \numprint{359706} \\
        Zobodat & \numprint{6231} & \numprint{92003} \\
        WP & \numprint{10000} & \numprint{176775} \\
        \midrule
        sum & \numprint{251624} & \numprint{5121436} \\
        \bottomrule
    \end{tabular}
    \caption{Number of tokens (incl. punctuation) and sentences within the evaluation samples.}
    \label{tab:number-sentences}
\end{table}

\subsection{Pipeline}
\label{sec:pipeline}

    Since HeidelTime requires single white spaces and, at least for some rules, part-of-speech information, all texts described in the previous section have been pre-processed using a TextImager pipeline \cite{Hemati:Uslu:Mehler:2016} as follows:
    \begin{enumerate}
        \item Normalization: All white spaces have been normalized to single spaces.
        
        \item Segmentation: Sentences have been segmented using the OpenNLP Max Entropy Model.\footnote{\texttt{opennlp-de-ud-gsd-sentence-1.0-1.9.3.bin}, Apache 2.0 License \url{https://opennlp.apache.org/index.html}}
        
        \item Tokenization: Word forms have been tokenized by using the Stanford CoreNLP \cite{Manning:2014} \textit{via} DKPro \cite{Eckart:2014}.
        
        \item Part-of-speech Tagging: Parts-of-speech (POS) have been assigned by using the POS tagger from MateTools \cite{Bohnet:2012} \textit{via}  DKPro \cite{Eckart:2014}
        
                    
    \end{enumerate}

    HeidelTime and \Name are run on the pre-processed texts and compared by means of two views in a UIMA CAS.

\subsection{Results}
\label{sec:results}

\Name found \numprint{4458} more \timex expressions than the original HeidelTime, a gain of \numprint[\%]{8.5}, as summarized in Tab.~\ref{tab:number-timex}.
However, as discussed below in Sec.~\ref{sec:discussion}, the bare gain in coverage has to put into perspective.

\begin{table}[htb]
    \centering
    \begin{tabular}{lll}
    \toprule
    Sample & HeidelTime & \Name \\
    \midrule
    Bundestag & \numprint{36068}  & \numprint{37507} \\
    DTA & \numprint{5827} & \numprint{7568} \\
    SZ & \numprint{4399} & \numprint{5305} \\
    Zobodat & \numprint{2110} & \numprint{2235} \\
    WP & \numprint{3966} & \numprint{4213} \\
    \midrule 
    sum & \numprint{52370} & \numprint{56828} \\
    \bottomrule
    \end{tabular}
    \caption{Number of \timex expressions found by original HeidelTime and \Name (see also Sec.~\protect\ref{sec:discussion}).}
    \label{tab:number-timex}
\end{table}

The coverage is detailed in Tab.~\ref{tab:coverage-details}.
Column \enquote{novel} lists the number of temporal expressions newly found by \Name, whereas the column \enquote{missing} counts the expressions only found by the original HeidelTime.
Since there are 298 missing \timex in total (assuming that these are true positives), this means that the newly added rules in \Name interfere with the application of some of the original rules.
The extended rule system of \Name also generally covers larger token spans, as expressed in the column \enquote{extended}.
Smaller token spans -- column \enquote{reduced} -- can be ignored due to their little frequency of occurrence.

\begin{table}[htb]
    \centering
    \begin{tabular}{lllll}
    \toprule
    Sample & novel & extended & reduced & missing \\ 
    \midrule
    Bundestag & \numprint{267} & \numprint{70} & 2 & \numprint{97} \\
    DTA & \numprint{258} & \numprint{37} & --- & \numprint{59} \\
    SZ & \numprint{436} & \numprint{75} & 1 & \numprint{103} \\
    Zobodat & \numprint{86} & \numprint{5} & --- & 8 \\
    WP & \numprint{133} & \numprint{28} & --- & \numprint{31} \\
    \midrule
    sum & \numprint{1180} & \numprint{215} & 3 & \numprint{298} \\
    \bottomrule
    \end{tabular}
    \caption{Coverage details. Novel: newly found temporal expressions; extended: \enquote{longer} matches; reduced: \enquote{shorter} matches; missing: failed detection by \Name.}
    \label{tab:coverage-details}
\end{table}

The following examples provide an impression of the kinds of temporal expressions which \Name is designed for: \textit{Vorjahr} \enquote*{preceding year}, \textit{übermorgen} \enquote*{the day after tomorrow}, \textit{Wintermonate} \enquote*{winter months}, \textit{eine halbe Stunde} \enquote*{half an hour}. 
Furthermore, durations as property denotations like \textit{dreitägig} \enquote*{three-day} as in \textit{dreitägige Exkursion} \enquote*{three-day excursion} are also taken to be \timex expressions. 
However, a manual inspection of the outcome of \Name revealed two kinds of expressions which need to be discussed.

\section{Discussion}
\label{sec:discussion}

To assess the outcome of \Name not only in quantitative terms, we manually inspected \numprint{1198} \timex expressions only found by \Name.\footnote{Bundestag: 220, Zobodat: 228, DTA: 312, WP: 155, SZ: 283.}
Each instance is labeled as being a true or a false positive -- see Tab.~\ref{tab:false-positives} for an overview.
The relatively high number of false positives, however, is mainly due to two kinds of overgeneralizations. 

The first overgeneralization is bound up with two \textsc{time} rules which wrongly apply not only to times of day expressions but to any measure terms formed with digits and a separating punctuation symbol.
From 168 examples of this type, which have been part of manual inspection, 139 are regarded false, and 29 true positives.
This overgeneralization is attenuated in the release of \Name as of  \printdate{2022-04-19} and following ones.

\begin{table}[htb]
    \centering
    \begin{tabular}{lll}
    \toprule
    Sample & true & false \\
    \midrule
    Bundestag & 185 & 35 \\
    DTA & 217 & 95 \\
    SZ & 210 & 73 \\
    Zobodat & 64 & 64 \\
    WP & 125 & 30 \\
    \midrule 
    sum & 801 & 297 \\
    \bottomrule
    \end{tabular}
    \caption{Instances of true and false positives in newly detected \timex expressions.}
    \label{tab:false-positives}
\end{table}

The second kind of overgeneralization is due to a significant proportion of newly detected occurrences of German \textit{nun}, which has not been part of the original HeidelTime.
\Name found \numprint{2530} instances in total.\footnote{Bundestag: \numprint{1010}, DTA: \numprint{1119}, SZ: \numprint{289}, Zobodat: 28, WP: 84.}
This particle word, however, has two broad uses which translate into English as \enquote*{now} respectively \enquote{well} (that is, the adverb typically used sentence initial).
Obviously, only the first of these uses is a temporal one, namely \textsc{present\_ref}.\footnote{Actually, things are more complicated than that: as is well known, even indexical expressions are evaluated with reference to three temporal indices, namely event time, reference time, and utterance time \cite{Reichenbach:1947}. For a semantics of temporal reference see \cite{Kamp:1979}.} 
During manual inspection, 455 instances of \textit{nun} (irrespective of capitalization) have been checked: 333 of them correspond to a temporal, 122 to a discourse use. 
Since there does not seem to be typical contexts which distinguish between temporal and discourse-functional \textit{nun}, overgeneration cannot simply be prevented by a negative rule. 
This example therefore exemplifies a limit of temponym recognition based on regular expressions.
The question therefore arises whether neural network-based approaches such as CNNs \cite{lin-etal-2017-representations}, which have been trained on date expressions, fare better with regard to temporal particles.\footnote{Neural networks can be used for temponym \emph{detection}, however, but not so easily for time \emph{normalization}.}
For the time being, a user of \Name can choose how to proceed by commenting out the rule in question (rule \texttt{date\_r8a-explicit}).

Removing the counts for \textit{nun} and for the overgeneralizing \textsc{time} rules from the figures given in Tab.~\ref{tab:number-timex}, we get the more \enquote{cautious} overview in Tab.~\ref{tab:number-timex-cleaned}.
There is now a gain of \numprint{1416} \timex expressions, or \numprint[\%]{2.7}.

\begin{table}[htb]
    \centering
    \begin{tabular}{lll}
    \toprule
    Sample & HeidelTime & \Name \\
    \midrule
    Bundestag & \numprint{36068}  & \numprint{36386} \\
    DTA & \numprint{5827} & \numprint{6429} \\
    SZ & \numprint{4399} & \numprint{4748} \\
    Zobodat & \numprint{2110} & \numprint{2136} \\
    WP & \numprint{3966} & \numprint{4087} \\
    \midrule 
    sum & \numprint{52370} & \numprint{53786} \\
    \bottomrule
    \end{tabular}
    \caption{Number of \timex expressions found by original HeidelTime and \Name with problematic cases removed.}
    \label{tab:number-timex-cleaned}
\end{table}

Note finally that we still found examples that circumvented our highly extended negative rules, cf. Sec.~\ref{subsec:negative-rules}.
For instance, the given name \textit{Carl}, as opposed to the form variant \textit{Karl}, is not part of any black list, so that the proper name \textit{Carl Winter} still triggers a season of the year rule.

\section{Conclusion}
\label{sec:conclusion}

Based on manually collected false negatives from various time banks and biological texts, we developed \Name, a German extension of HeidelTime.
We constructed an evaluation corpus to quantify the gain in coverage of the extension.
A manual inspection of novel instances found by \Name identified two kinds of rules which tend to overgeneralize to non-temporal instances, however.
Depending on how much potential overgeneralization is admitted, the relative coverage improvement of \Name is between \numprint[\%]{2.7} and \numprint[\%]{8.5}.
At this point, one could object that our approach mainly discovers rarely occurring expressions of time. 
But especially these cases are interesting for disciplines of the humanities or biodiversity (as outlined in Secs.~\ref{sec:motivation} and \ref{subsec:manual-exploration}), which also deal with what is rare rather than with what is frequent.

Future work might deal with a compositional approach to written numbers such as \textit{einhundertdreiundfünfzig} \enquote*{one hundred and fifty-three}.
A difficulty here of course is to generate a corresponding norm value.
Another issue is the improvement of negative rules to prevent overgeneralization.
We used a list-based approach, making use of governmental material and material generated by a BERT model. 
\Name is  available  from \url{https://github.com/texttechnologylab/heideltime}.

\section{Acknowledgements}

This work is funded by the DFG, grant ME2746/5-2.

\section{Bibliographical References}
\label{reference}

\bibliographystyle{lrec}
\bibliography{heidelbib2020}

\section{Language Resource References}
\label{lr:ref}
\bibliographystylelanguageresource{lrec}
\bibliographylanguageresource{languageresource}

\end{document}